\documentclass{egpubl}
\usepackage{eg2027}
\renewcommand{\today}{September 2026}
\preprint
\copyrightTextTitPag{}
\copyrightTextRunPag{}

\usepackage[T1]{fontenc}
\usepackage{mathptmx}
\usepackage{cite}
\BibtexOrBiblatex
\electronicVersion
\PrintedOrElectronic
\usepackage{graphicx}
\usepackage{egweblnk}

\usepackage{amsmath}
\usepackage{amssymb}
\usepackage{booktabs}
\usepackage{placeins}

\providecommand{\Description}[1]{}
\newcommand{\tablefont}{\scriptsize}

\title[Geometry-Preserving Point Cloud Watermarking]{Geometry-Preserving Blind Watermarking for Raw 3D Point Clouds}
\author[Rungui Zhou et al.]{Rungui Zhou \quad Chuanzhi Zhou \quad Ruihuan Wang \quad Peng-Shuai Wang\\Peking University}
\hypersetup{pdfsubject={Preprint},pdfkeywords={point clouds, blind watermarking, geometry preservation}}

\begin{document}

\maketitle
\hypersetup{pdfauthor={Rungui Zhou, Chuanzhi Zhou, Ruihuan Wang, Peng-Shuai Wang},pdftitle={Geometry-Preserving Blind Watermarking for Raw 3D Point Clouds}}

\begin{abstract}
Raw 3D point clouds are a core geometric representation. Establishing their ownership is challenging because point sets are irregular, unstructured, and frequently altered by resampling and geometric preprocessing. We present a blind watermarking framework that operates directly on xyz coordinates and supports both object-level shapes and scene-scale scans. At verification time, the embedded message is recovered from the observed point cloud alone, without access to the original point cloud, color, normals, or mesh connectivity. The method jointly learns watermark embedding and extraction through a feed-forward octree-based architecture, enabling efficient multi-scale geometric reasoning on large point sets. During training, a stochastic transformation layer exposes the decoder to common geometric perturbations, while progressive pose alignment improves robustness to pose changes.
Experiments on object-level and scene-level benchmarks demonstrate reliable message recovery under common geometric processing while maintaining low geometric distortion. Qualitative comparisons further show that the learned perturbations are less visually conspicuous and less spatially structured than those of handcrafted alternatives.

\end{abstract}

\section{Introduction}
\label{sec:intro}

3D point clouds are a core representation for geometry processing, spatial capture, robotics, and digital twins~\cite{guo2020deep}. They are routinely acquired, exchanged, and reprocessed for tasks such as reconstruction, mapping, and simulation. Once a point set leaves the original acquisition pipeline, verifying its provenance or ownership becomes important. Watermarking offers a way to address this problem by encoding a message directly in the geometry.
Yet blind watermarking of raw point clouds remains challenging, especially when verification must succeed after realistic geometric processing.
The embedded signal must be spread across the point cloud to survive changes during processing.
At the same time, it must remain subtle enough to avoid visible surface patterns.
\looseness=-1

The difficulty stems from the nature of raw point-cloud data. Point sets are unordered, irregularly sampled, and often contain only geometry, without color, normals, or mesh connectivity. In practical content pipelines, they are often affected by acquisition noise, cropping, voxelization, and smoothing~\cite{zhou2018open3d,rusu20113d}. These operations change both point coordinates and the local neighborhoods that many geometry-based methods rely on. These constraints motivate an xyz-only blind watermarking setting. In deployment, the verifier may receive only a redistributed or transformed copy, with no access to the original point cloud. The verifier must therefore recover the message from the observed point cloud alone, without the original geometry or auxiliary attributes.

Existing 3D watermarking methods face several limitations in this setting. Many rely on mesh connectivity~\cite{wang2022deep3Dmesh} or appearance-rich neural 3D representations~\cite{kerbl20233guassian}, neither of which is directly available in raw point clouds. Geometry-only approaches often use hand-crafted geometric modulation or hybrid pipelines that retain manually designed embedding rules~\cite{liu2018blindpctrad1,liu2019novelpctrad2,zaman2025deeppcd}.
Watermarking of scene-scale raw scans also remains less explored. These scans routinely undergo cropping, voxelization, smoothing, and resampling during storage, registration, mapping, and preprocessing, making robustness to geometric processing a practical requirement.

In this work, we propose an end-to-end neural framework for blind watermarking of raw point clouds using xyz coordinates alone. The framework jointly learns watermark embedding and blind extraction in a single feed-forward pipeline, without hand-crafted carriers or test-time optimization. It requires no RGB, normals, or mesh connectivity. To handle irregular density and large point counts, we adopt an octree-based neural backbone for efficient multi-scale processing~\cite{wang2017ocnn}. The hierarchy efficiently combines local and global geometric context. This allows the same architecture to scale from object-level point clouds to large scene scans. We train separate models to account for differences between object and scene data, while keeping the architecture unchanged.
\looseness=-1

To improve robustness to geometric transformations, we use a stochastic simulation layer that applies noise, smoothing, cropping, rotation, scaling, and voxelization during training.
We also use progressive pose alignment before octree decoding to account for changes in global pose.
To preserve local geometry, we use local covariance matching to encourage nearby points to retain a similar spatial arrangement after embedding. We further encourage neighboring points to move by similar amounts and in similar directions.
This displacement smoothness helps reduce local surface artifacts.
These geometry-preserving objectives help preserve visual fidelity, while attack simulation improves message recovery after processing.
\looseness=-1

In summary, our contributions are threefold. First, we formulate an xyz-only blind watermarking pipeline for raw point clouds that performs feed-forward embedding and extraction across irregular point sets. Second, we develop a geometry-preserving training formulation that combines transformation simulation and pose alignment for robustness with local structure constraints for embedding fidelity. Third, we evaluate this unified design on object-level shapes and scene-scale scans, including statistical, computational, and qualitative analyses under common point-cloud processing operations.
The resulting verifier requires neither a fixed point ordering nor point-to-point correspondence with a stored reference. Both can be lost after resampling or voxel filtering. The verifier can therefore operate on redistributed copies with an unknown sampling history.
Our code is available in the \href{https://anonymous.4open.science/r/watermark_release-47F8/}{project repository}.
\looseness=-1

\section{Related Work}
\label{sec:related_work}

\subsection{Robust Neural Watermarking}
Deep watermarking introduced encoder--decoder formulations in which message embedding and extraction are learned jointly, often with differentiable distortion layers that expose the decoder to anticipated transformations during training~\cite{zhu2018hidden,mareen2024blind,tancik2020stegastamp}. This line of work establishes the importance of jointly balancing payload, recoverability, and imperceptibility. Raw point clouds present a distinctly different carrier: they are unordered, irregularly sampled, and have no fixed raster neighborhoods. Consequently, the geometric transformations relevant to point clouds---such as resampling, voxelization, and partial observation---must be modeled directly, and methods designed for dense 2D grids do not transfer unchanged.

\subsection{Watermarking of Geometric Representations}
3D watermarking has a long history in meshes and other structured geometric models. Classical methods typically embed information in spatial or spectral domains by exploiting surface connectivity, well-defined neighborhoods, and differential properties such as normals and curvature~\cite{wang2008comprehensivemeshsurvey,praun1999robust,tang2023dualmesh}. More recent learning-based approaches extend this line with graph- and mesh-based neural operators trained for robustness~\cite{wang2022deep3Dmesh,zhu2025mesh}. These methods, however, fundamentally assume access to mesh connectivity or stable topology. Such assumptions do not hold for raw point clouds, particularly for large-scale scanned scenes acquired directly as unstructured point sets. Converting scans into meshes through reconstruction may introduce geometric artifacts, alter local structure, and confound evaluation. Consequently, mesh-based watermarking methods are not directly comparable to our geometry-only point-cloud setting.

\subsection{Watermarking for Point Clouds}
Point-cloud watermarking remains comparatively underexplored despite the widespread use of LiDAR and RGB-D scanning. Early methods are largely blind and hand-crafted. Some rely on appearance channels such as color~\cite{ferreira2020robust_color,zhang2024robust_zeromark}, while geometry-only approaches embed bits by modulating geometric statistics, such as spherical or angular encodings, distance distributions, or local descriptors chosen for invariance to rigid transformations~\cite{liu2018blindpctrad1,liu2019novelpctrad2,wu2024comprehensive,ke2013self}. A common design pattern is to first construct a handcrafted reference frame or carrier set from relatively stable geometric cues, for example through PCA-aligned coordinates, curvature-based point selection, or radial partitioning, and then encode bits by modulating distances or angles within that anchored structure. These methods are often interpretable and can perform well under controlled distortions, but their robustness is tightly tied to the stability of the underlying geometric anchors under resampling and local structural change.

This dependence becomes problematic in modern scanning pipelines, where raw point sets are repeatedly subsampled, voxelized, merged, and filtered before downstream use. Under such structural perturbations, including aggressive resampling, heavy cropping, and severe noise, hand-crafted carriers can become unstable and decoding performance may degrade substantially. A more recent line of work introduces deep networks primarily for blind extraction while retaining a manually designed embedding transform, for example through block-wise manipulation~\cite{zaman2025deeppcd}. This hybrid strategy improves robustness but does not jointly learn embedding and extraction end-to-end.

Our approach differs in two respects. First, the framework learns both watermark embedding and blind extraction jointly through feed-forward inference, without hand-crafted embedding rules or test-time optimization. Second, it explicitly models preprocessing operations such as voxel quantization, which are ubiquitous in geometric pipelines but not always emphasized in prior evaluations. Existing studies also overwhelmingly focus on object-level benchmarks~\cite{chang2015shapenet,wu20153dmodelnet}, whereas large raw scans, such as ScanNet-style scenes~\cite{dai2017scannet,yeshwanth2023scannet++}, remain largely absent. We address this gap by applying the same unified architecture and training recipe to both object-level and scene-level data, with separate training only to account for distribution shift.

\subsection{Watermarking for Neural 3D Representations}
Recent neural 3D representations have also motivated watermarking methods for NeRF~\cite{mildenhall2021nerf} and 3D Gaussian Splatting (3DGS)~\cite{kerbl20233guassian}~\cite{luo2023copyrnerf,li2023steganerf,huang2024gaussianmarker,zhao2025rdsplat,huang2025marksplatter}. Many of these approaches embed messages into appearance-related parameters, such as color, opacity, or spherical harmonics, and often rely on per-scene optimization or fine-tuning to preserve rendering quality~\cite{huang2024gaussianmarker}. Our focus is different: we operate directly on raw xyz geometry, require neither mesh connectivity nor appearance attributes, and use efficient feed-forward embedding and extraction. This makes the method suitable for large scanned point sets and preprocessing-heavy geometric workflows in which operations such as voxelization are routine.

\section{Method}
\label{sec:method}

\begin{figure*}[t]
  \centering
  \includegraphics[width=\textwidth]{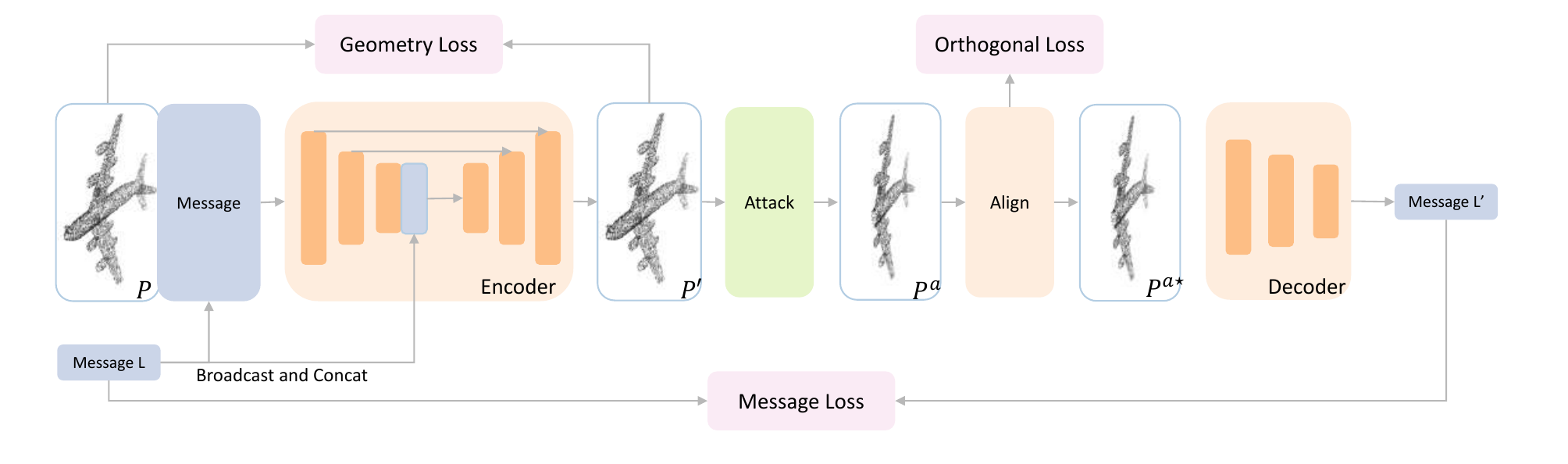}
  \caption{Overview of our framework. $P$ is the normalized input point cloud, $m\in\{0,1\}^{L}$ is the target $L$-bit message, and $P'=E_{\theta}(P,m)$ is the watermarked point cloud. During training, the attack module produces $P^{a}=\mathcal{A}(P')$; two learned transforms then map it to the pose-aligned cloud $P^{a\star}$. The decoder predicts the recovered message $\hat m$ from $P^{a\star}$ alone. Geometry loss compares $P$ with $P'$, message loss compares $m$ with the decoder prediction, and orthogonal loss regularizes the two alignment transforms.}
  \Description{An architecture diagram showing an octree-based encoder, a stochastic attack module, two alignment stages in the decoder, and final blind watermark extraction.}
  \label{fig:architecture}
\end{figure*}

Given an input point cloud $P=\{p_i\}_{i=1}^{N}$ with $p_i\in\mathbb{R}^3$ and a binary message $m\in\{0,1\}^{L}$, our goal is to produce a watermarked point cloud $P'=\{p'_i\}_{i=1}^{N}$ that preserves the input geometry while allowing the message to be recovered after common geometric distortions. We normalize the raw coordinates to a canonical unit cube before processing and use $P$ hereafter for this normalized input.
Recovery is blind: the decoder receives only the distorted observation $P^{a}$, without access to the original point cloud.
\looseness=-1

\subsection{Architecture}
Our framework jointly trains an encoder and a decoder end-to-end. Given $P$ and $m$, the encoder produces
\begin{equation}
P'=E_{\theta}(P,m).
\end{equation}
From the attacked point cloud $P^{a}=\mathcal{A}(P')$, the decoder predicts message logits $\ell$ and recovers the bits by thresholding their sigmoid probabilities:
\begin{equation}
\ell=D_{\phi}(P^{a})\in\mathbb{R}^{L},\qquad \hat{m}=\mathbf{1}[\sigma(\ell)>0.5].
\end{equation}
Here $\mathcal{A}(\cdot)$ denotes the attack module, $\mathcal{N}(\cdot)$ denotes the unit-cube normalization used again inside the alignment module, and $\sigma(\cdot)$ is the sigmoid function.

\textbf{Encoder.}
The encoder is built on an octree-based U-Net with skip connections~\cite{wang2017ocnn}. Given the normalized input point cloud $P$, we construct an octree at a prescribed depth and use the local displacement at the finest level as the input signal. To increase embedding capacity, we inject the message $m$ at both the input stage and the bottleneck by broadcasting its bits as per-node features. Let $F_d$ denote the features at octree depth $d$ and $B$ the bottleneck features. We concatenate the message with each:
\[
  F'_d=[F_d\,\|\,\operatorname{broadcast}(m)],
  \qquad
  B'=[B\,\|\,\operatorname{broadcast}(m)].
\]
This conditions the encoder on the message at both stages. The octree hierarchy captures multi-scale geometric context while remaining practical for dense point sets.

\textbf{Attack simulation.}
During training, a stochastic attack module $\mathcal{A}(\cdot)$ perturbs the watermarked point cloud before decoding. The simulated attacks include additive Gaussian noise, smoothing, anisotropic stretching, random point dropout, random spherical cropping, random rotation, and random voxel downsampling. These operations model common non-adaptive processing; they do not explicitly optimize for watermark removal. For voxel downsampling, we follow the voxel-grid filtering strategy in Open3D~\cite{zhou2018open3d}: we partition space into a $2^D \times 2^D \times 2^D$ grid and replace the points in each occupied voxel with their centroid:
\[
\mathcal{V}_D(P')
=
\left\{
\frac{1}{|S_v|}\sum_{p'_n\in S_v} p'_n
\;\middle|\;
S_v\neq\varnothing
\right\}.
\]
Here $S_v=\{p'_n\in P':\lfloor 2^D p'_n\rfloor=v\}$ is the set of watermarked points assigned to voxel $v$.
For rotation, we gradually widen the angle range during training to stabilize optimization and help avoid early collapse.

\textbf{Decoder.}
The decoder consists of a lightweight progressive alignment module followed by an octree-based ResNet classifier.
Inspired by spatial transformer networks~\cite{jaderberg2015spatial_tnet}, each alignment head is a PointNet-style transform regressor~\cite{qi2017pointnet}: shared pointwise multilayer perceptrons map the xyz coordinates to per-point features, symmetric max pooling aggregates them into a permutation-invariant global descriptor, and fully connected layers regress a $3\times3$ matrix initialized around the identity. Cascading two such heads lets the first correct the dominant pose error and the second refine the residual misalignment before octree construction.
The first head takes the normalized attacked cloud $\mathcal{N}(P^{a})$ and predicts a transformation matrix $R_1$ for coarse alignment. The second head refines this alignment with a transformation $R_2$. We then renormalize the aligned cloud:
\begin{equation}
    \begin{aligned}
        P^{a1} &= R_1 \big( \mathcal{N}(P^{a}) \big), \\
        P^{a\star} &= \mathcal{N} \big( R_2 (P^{a1}) \big),
    \end{aligned}
    \label{eq:alignment}
\end{equation}
where $P^{a\star}$ is the aligned point cloud in canonical space. The extractor processes this representation to predict the message logits:
\begin{equation}
    \ell = D_{\phi}(P^{a\star}).
\end{equation}
Aligning the cloud before octree construction improves decoding stability under large geometric transformations.

\subsection{Loss Functions}
We train the framework with a weighted sum of a decoding loss $\mathcal{L}_{\text{bce}}$, a geometry-preserving loss $\mathcal{L}_{\text{geo}}$, and an orthogonality regularizer on the two alignment matrices $\mathcal{L}_{\text{orth}}$:
\begin{equation}
\mathcal{L} = \lambda_{\text{dec}}\mathcal{L}_{\text{bce}}(\ell, m)
            + \lambda_{\text{geo}}\mathcal{L}_{\text{geo}}(P', P)
            + \lambda_{\text{orth}}\mathcal{L}_{\text{orth}}(R_1, R_2) .
\end{equation}

For the decoding loss, we use per-bit binary cross-entropy $\mathcal{L}_{\text{bce}}$ between the predicted logits $\ell$ and the target message $m$.
For the orthogonality regularizer, we encourage the two predicted alignment matrices $(R_1, R_2)$ in Eq.~\ref{eq:alignment} to be orthogonal:
\begin{equation}
\mathcal{L}_{\text{orth}}(R_1, R_2)=\|R_1R_1^\top-I\|_F^2+\|R_2R_2^\top-I\|_F^2.
\end{equation}

For the geometry-preserving loss, we combine a symmetric Chamfer distance term $\mathcal{L}_{\text{cd}}$ with two structural regularizers:
\begin{equation}
  \mathcal{L}_{\text{geo}} = \mathcal{L}_{\text{cd}} + \beta\,\mathcal{L}_{\text{struct}} + \gamma\,\mathcal{L}_{\text{smooth}},
\end{equation}
where $\lambda_{\text{geo}}=1$,
$\mathcal{L}_{\text{struct}}$ is designed to preserve the spatial arrangement of nearby points during watermark embedding,
and $\mathcal{L}_{\text{smooth}}$ encourages neighboring points to move by similar amounts and in similar directions.
\looseness=-1

Specifically, we match local second-order statistics between the original and watermarked clouds. This follows the use of second-order statistics to characterize structure in PCA-based local geometry analysis~\cite{hoppe1992surfacePCA} and Gram-matrix representations~\cite{gatys2016image}.
Let $\mathcal{N}_k(i)$ denote the indices of the $k$-nearest neighbors ($k$-NN) of point $p_i$ in the original cloud $P$. We use the same neighbor indices in the watermarked cloud $P'$ to compare corresponding neighborhoods before and after embedding. For $X\in\{P,P'\}$, the local covariance matrix is
\[
C_i(X)=\frac{1}{k}\sum_{j\in\mathcal{N}_k(i)}(x_j-\bar x_i)(x_j-\bar x_i)^\top,
\quad
\bar x_i=\frac{1}{k}\sum_{j\in\mathcal{N}_k(i)}x_j .
\]
We then penalize the covariance difference within each corresponding neighborhood:
\begin{equation}
  \mathcal{L}_{\text{struct}}(P,P')=\frac{1}{N}\sum_{i=1}^{N}\left\|C_i(P')-C_i(P)\right\|_F .
\end{equation}
Using the same $k$-NN graph provides consistent neighborhoods for comparison and avoids recomputing the graph on $P'$ during loss evaluation.

To complement covariance matching, which constrains local geometry, we regularize the spatial variation of the embedding displacements~\cite{sorkine2005laplacian}. Let $d_i=p'_i-p_i$ denote the displacement of point $i$. Using the same neighborhood graph, we encourage neighboring points to undergo similar displacements:
\begin{equation}
  \mathcal{L}_{\text{smooth}}
  =\frac{1}{N}\sum_{i=1}^{N}\frac{1}{k}\sum_{j\in\mathcal{N}_k(i)}\left\|d_i-d_j\right\|_2^2 .
\end{equation}
This term penalizes high-frequency variations in the displacement field and helps suppress local surface artifacts. Together, the two regularizers encourage the embedding to preserve local geometry through spatially coherent point displacements.

\setlength{\textfloatsep}{12pt plus 2pt minus 2pt}
\setlength{\floatsep}{9pt plus 2pt minus 2pt}
\setlength{\intextsep}{12pt plus 2pt minus 2pt}
\setlength{\dbltextfloatsep}{12pt plus 2pt minus 2pt}
\section{Experiments}\label{sec:experiments}

\subsection{Datasets}
\textbf{Object-level point clouds.}
We evaluate our method on ModelNet40~\cite{wu20153dmodelnet} for direct comparison with recent baselines and on ShapeNet-style objects~\cite{chang2015shapenet} to assess scalability beyond small, fixed-size inputs. Our framework supports variable-size point clouds. For ShapeNet, we randomly sample $N\in[4000,10000]$ points per shape during training and use an octree depth of $d=5$. For ModelNet40, we follow the matched evaluation protocol with a fixed input size of $N=1024$ and octree depth $d=5$.

\textbf{Scene-level point clouds.}
To assess performance at scene scale, we use ScanNet~\cite{dai2017scannet} for indoor scans and SemanticKITTI~\cite{behley2019semantickitti} for outdoor LiDAR scenes, covering different sensing conditions and scene statistics. For both datasets, we use $N\in[20000,100000]$ points and an octree depth of $d=7$. For the Gaussian noise attack, we use a smaller standard deviation of $\sigma=0.001$ because dense scene scans are more sensitive to local structural perturbations. Following the official SemanticKITTI split, we train on sequences 00--10 and evaluate on sequences 11--21. For computational efficiency, we randomly sample 100 scans from each training and test sequence.

\subsection{Evaluation Protocol}
\textbf{Message setting.}
For each point cloud, we embed a random binary message $m\in\{0,1\}^{L}$. Unless otherwise noted, we use a payload of $L=16$ bits for ShapeNet and the scene-level datasets. ModelNet40 follows the matched 2-bit protocol used by the recent object-level baseline. We measure robustness using bitwise accuracy (BitAcc) and report the message-level success rate, defined as the fraction of messages for which all bits are recovered correctly.

\textbf{Attacks.}
We evaluate robustness against a broad set of common distortions: Gaussian noise (additive i.i.d.\ noise with value clipping, using $\sigma=0.01$ for objects and $\sigma=0.001$ for scenes), $k$NN smoothing (replacing each point with the centroid of itself and its $k=16$ nearest neighbors), random anisotropic stretching, random point dropout (retaining 70\% of points), random spherical cropping~\cite{lai2022spherecrop}, random rotation (typically around the $x$-axis with $\theta\in[-\alpha,\alpha]$), and voxel downsampling (quantizing points into a $2^d\times 2^d\times 2^d$ grid and replacing each occupied voxel with its centroid). Unless explicitly listed as a combined condition, these transformations are evaluated independently. This protocol characterizes robustness to common non-adaptive processing; it does not by itself establish resistance to an adaptive remover that optimizes directly against the watermark.

\textbf{Geometry quality.}
We assess geometric fidelity using the symmetric Chamfer distance (CD), peak signal-to-noise ratio (PSNR), and normal consistency (NC) between the original and watermarked point clouds. These metrics are reported only for the \emph{clean} watermarked outputs. Because CD and PSNR are global measures, spatially coherent local artifacts may remain visible even when their contribution to the global error is small.
\looseness=-1

\subsection{Implementation Details}
Unless otherwise specified, we train all models end-to-end using AdamW with a cosine learning-rate schedule, an initial learning rate of $\texttt{1e-3}$, and zero weight decay. The batch size is $\texttt{8}$ for scene-level scans and $\texttt{24}$ for object-level shapes. We fix the decoding loss weight at $\lambda_{\text{dec}}=\texttt{1.0}$, set $\lambda_{\text{geo}}=1$, use $k=20$ neighbors for the geometry terms, and linearly ramp the geometry and local regularizer weights during training. The alignment heads use an orthogonality penalty with weight $\texttt{0.01}$. The rotation curriculum increases the maximum rotation angle every 15 epochs until it reaches the dataset-specific limit. To support size-agnostic inference, we train the ShapeNet and scene-level models with variable-size point clouds as described above. ModelNet40 follows the fixed-size matched protocol. All point clouds are normalized to the unit cube before octree construction.

\textbf{Statistical and computational reporting.}
For each dataset, we report the mean and standard deviation across the evaluated point clouds using a fixed evaluation seed. These standard deviations describe variation across samples, not across independent training runs. Table~\ref{tab:reliability_cost} summarizes clean-watermark fidelity, clean decoding accuracy, end-to-end inference time, and peak allocated GPU memory on an NVIDIA RTX 4090. Inference time includes embedding and clean decoding for one sample.

\begin{table}[t]
  \centering
  \caption{Sample-level reliability and computational cost. Values are mean $\pm$ standard deviation over test point clouds unless stated otherwise. PSNR is in dB; time (ms) and memory (GB) are measured per sample on an NVIDIA RTX 4090.}
  \label{tab:reliability_cost}
  \tablefont
  \setlength{\tabcolsep}{7pt}
  \begin{tabular}{@{}lccccc@{}}
    \toprule
    Dataset & BitAcc & CD ($10^{-4}$) & PSNR & Time & Mem. \\
    \midrule
    ModelNet40 & $0.996\pm0.047$ & $10.7\pm6.6$ & $37.17\pm2.80$ & 44.92 & 0.44 \\
    ShapeNet & $0.963\pm0.078$ & $1.7\pm1.03$ & $45.2\pm2.67$ & 45.85 & 0.71 \\
    ScanNet & $0.977\pm0.050$ & $0.617\pm0.215$ & $48.88\pm1.55$ & 66.17 & 1.25 \\
    SemanticKITTI & $0.999\pm0.012$ & $1.08\pm0.44$ & $43.61\pm1.04$ & 68.45 & 1.64 \\
    \bottomrule
  \end{tabular}
\end{table}

\begin{figure}[!tbh]
  \centering
  \includegraphics[width=0.82\columnwidth]{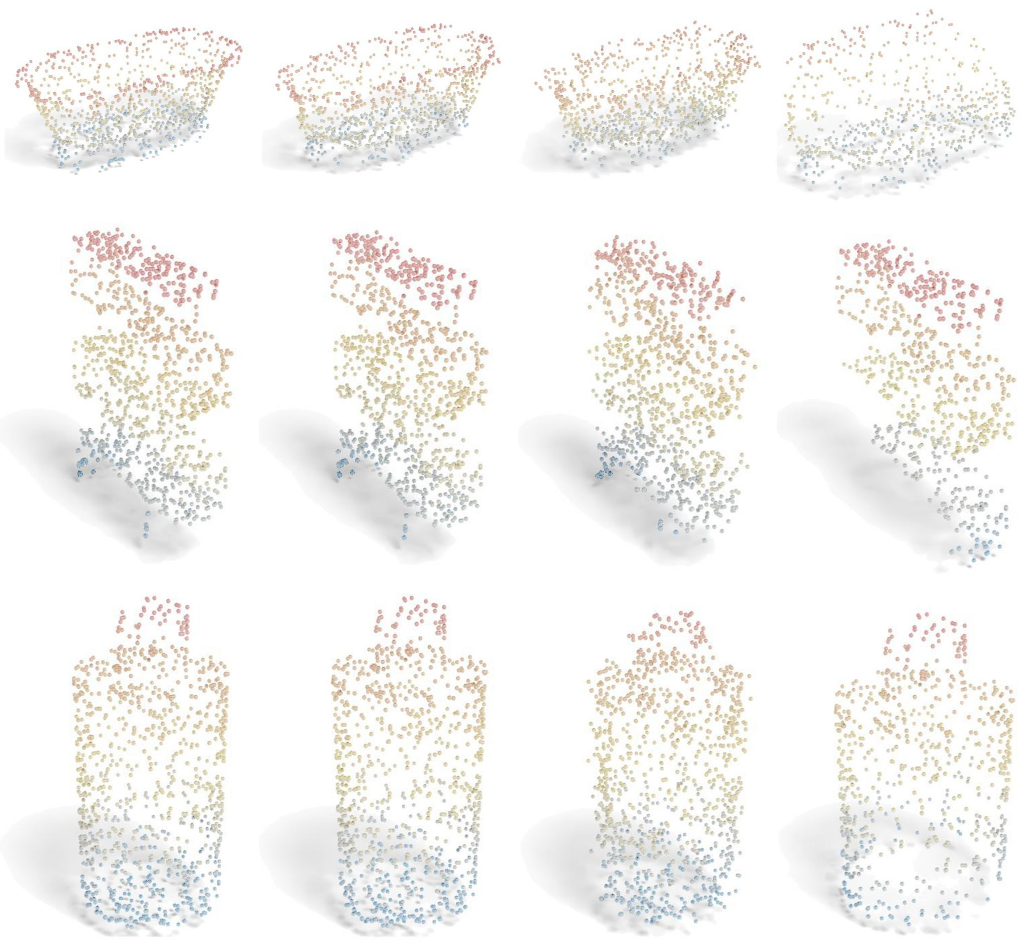}
  \caption{ModelNet40 examples. Columns are Original, Watermarked, Noise, and Crop; each row shows the same object.}
  \label{fig:qual_modelnet_main}
\end{figure}

\subsection{Quantitative Results}
\begin{figure*}[t]
  \centering
  \begin{minipage}{\textwidth}
    \centering
    \small
    \makebox[0.25\linewidth][c]{Original}%
    \makebox[0.25\linewidth][c]{Ours}%
    \makebox[0.25\linewidth][c]{\cite{feng2015distance}}%
    \makebox[0.25\linewidth][c]{\cite{liu2019novelpctrad2}}\\[-0.4ex]
    \includegraphics[width=\linewidth]{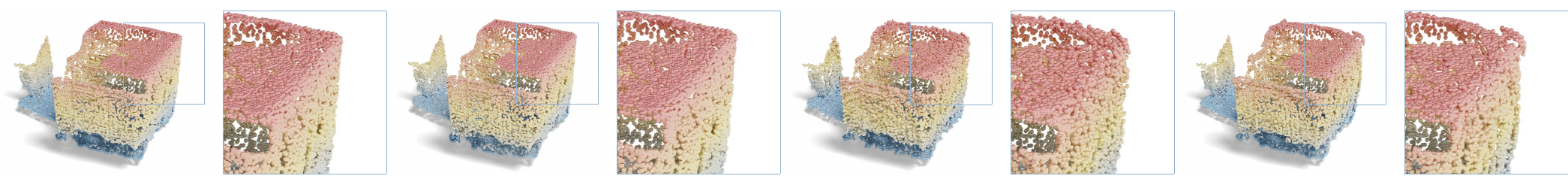}
  \end{minipage}
  \caption{ScanNet comparison. Each group pairs a full view with a close-up of the blue-boxed region. Ours better preserves local surface appearance; the handcrafted baselines~\cite{feng2015distance,liu2019novelpctrad2} exhibit coherent shell-like artifacts.}
  \label{fig:qual_compare_main}
\end{figure*}

\textbf{Comparison on ModelNet40.}
We first compare our method against recent baselines on ModelNet40. For a matched comparison with AETiC'25~\cite{zaman2025deeppcd}, which uses a hybrid SVD and PointNet++ pipeline, we follow the same protocol: $N=1024$, a 2-bit payload, and identical attack settings. As shown in Table~\ref{tab:vertical_compare}, our method achieves higher decoding accuracy under clean conditions and every listed attack, while reducing the Chamfer distance by roughly two orders of magnitude (0.001 vs.\ 0.105). These results show improvements in both robustness and geometric fidelity.

\begin{table}[t]
  \centering
  \caption{ModelNet40 comparison with 1024 points and a 2-bit payload. Baseline results are from AETiC'25, and Ours (2b) follows the same protocol. Our method achieves higher BitAcc under clean conditions and all listed attacks, with lower CD and higher PSNR.}\label{tab:vertical_compare}
  \tablefont
  \setlength{\tabcolsep}{18pt}
  \begin{tabular}{@{}lccc@{}}
    \toprule
    Setting / Attack & SVD  & PN++  & Ours (2b) \\
    \midrule
    Dataset & \multicolumn{3}{c}{ModelNet40} \\

    \#Points (train/test) & \multicolumn{3}{c}{1024} \\

    Watermark payload (bits) & \multicolumn{3}{c}{2} \\

    Chamfer Distance (CD) & 0.105 & 0.105 & \textbf{0.001}\\
    PSNR (dB) &  21.9 &  21.9 & \textbf{37.2}\\
    \midrule
    Clean              & 0.950 & 0.950 & \textbf{0.996} \\
    Noise ($\sigma=0.01$)                 & 0.950 & 0.778 & \textbf{0.994} \\
    Noise ($\sigma=0.03$)                 & 0.903 & 0.777 & \textbf{0.989}\\
    Smoothing                      & 0.957 & 0.795 & \textbf{0.991} \\
    Rotation ($[0, \pi]$)              & 0.955 & 0.633 & \textbf{0.959} \\
    Quantization                   & 0.955 & 0.788 & \textbf{0.990} \\
    Affine                         & 0.867 & 0.773 & \textbf{0.991} \\
    Shuffle                        & 0.588 & 0.780 & \textbf{0.990} \\
    Cropping (0.7)                 & 0.580 & 0.818 & \textbf{0.957} \\
    Dropout                        & 0.572 & 0.820 & \textbf{0.990}\\
    Noise + Dropout                & 0.572 & 0.810 & \textbf{0.987}\\
    \bottomrule
  \end{tabular}
\end{table}

\textbf{Large-scale evaluations.}
Beyond the small, fixed-size ModelNet40 inputs, we evaluate ShapeNet, ScanNet, and SemanticKITTI with a 16-bit payload (Table~\ref{tab:main_large_scale}). Our method maintains high decoding accuracy under the evaluated distortions with CD on the order of $10^{-4}$. On ShapeNet, CD is $1.7\times10^{-4}$ and PSNR is 45.2\,dB; Liu~\cite{liu2019novelpctrad2} has better global fidelity, while our method achieves higher BitAcc under every listed condition.

We compare against Feng~\cite{feng2015distance} and Liu~\cite{liu2019novelpctrad2}. Feng segments a point cloud into patches, constructs a PCA-based spherical coordinate system, and embeds one bit per patch through AQIM-based angle modulation. Liu selects high-curvature vertices as carriers, constructs a synchronization frame from low-curvature vertices, partitions the cloud into radial ball rings, and embeds bits by modulating the radii of candidate vertices.

\begin{table}[t]
  \centering
  \caption{Geometric fidelity and decoding accuracy on ShapeNet, ScanNet, and SemanticKITTI with a 16-bit payload. Attack rows report BitAcc; PSNR and CD are measured on clean watermarked outputs.}\label{tab:main_large_scale}
  \tablefont
  \setlength{\tabcolsep}{2pt}
  \begin{tabular}{@{}lccccccccc@{}}
    \toprule
    & \multicolumn{3}{c}{ShapeNet} & \multicolumn{3}{c}{ScanNet} & \multicolumn{3}{c}{SemKITTI} \\
    \cmidrule(lr){2-4} \cmidrule(lr){5-7} \cmidrule(lr){8-10}
     & Ours & \cite{liu2019novelpctrad2} & \cite{feng2015distance} & Ours & \cite{liu2019novelpctrad2} & \cite{feng2015distance} & Ours & \cite{liu2019novelpctrad2} & \cite{feng2015distance} \\
    \midrule
    PSNR   & 45.2 & \textbf{47.6} & 45.2 & \textbf{49.0} & 43.4 & 38.7 & \textbf{43.6} & 42.2 & 42.8 \\
    CD     & 1.7e-4 & \textbf{1.5e-4} & 1.7e-4 & \textbf{6.1e-5} & 1.0e-4 & 3.0e-4 & \textbf{1.0e-4} & 1.2e-4 & 1.2e-4 \\
    \midrule
    Clean   & \textbf{0.963} & 0.812 & 0.734 & \textbf{0.977} & 0.778 & 0.657 & \textbf{0.999} & 0.488 & 0.833 \\
    Noise   & \textbf{0.944} & 0.600 & 0.537 & \textbf{0.971} & 0.734 & 0.650 & \textbf{0.995} & 0.471 & 0.757 \\
    Smooth  & \textbf{0.962} & 0.810 & 0.722 & \textbf{0.970} & 0.771 & 0.652 & \textbf{0.991} & 0.523 & 0.828 \\
    Stretch & \textbf{0.948} & 0.689 & 0.577 & \textbf{0.973} & 0.616 & 0.567 & \textbf{0.997} & 0.501 & 0.630 \\
    Dropout & \textbf{0.951} & 0.666 & 0.600 & \textbf{0.965} & 0.599 & 0.633 & \textbf{0.992} & 0.498 & 0.761 \\
    Crop    & \textbf{0.929} & 0.534 & 0.527 & \textbf{0.927} & 0.503 & 0.537 & \textbf{0.985} & 0.496 & 0.547 \\
    Rot.    & \textbf{0.907} & 0.812 & 0.732 & \textbf{0.888} & 0.778 & 0.650 & \textbf{0.978} & 0.510 & 0.832 \\
    Voxel   & \textbf{0.945} & 0.490 & 0.522 & \textbf{0.967} & 0.573 & 0.542 & \textbf{0.989} & 0.494 & 0.509 \\
    \bottomrule
  \end{tabular}
\end{table}

\begin{figure}[!tbh]
  \centering
  \includegraphics[width=1.0\columnwidth]{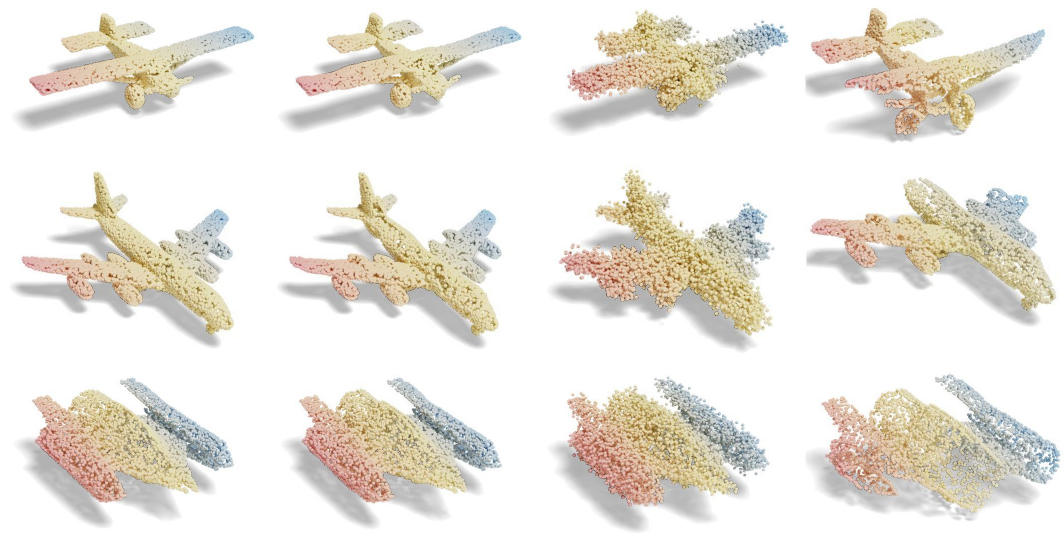}
  \caption{ShapeNet examples: Original, Watermarked, Noise, and Crop. Embedding preserves the geometry, while severe attacks still allow high decoding accuracy (Table~\ref{tab:main_large_scale}).}
  \label{fig:qual_shape_main}
\end{figure}

\begin{figure}[!tbh]
  \centering
  \includegraphics[width=0.94\columnwidth]{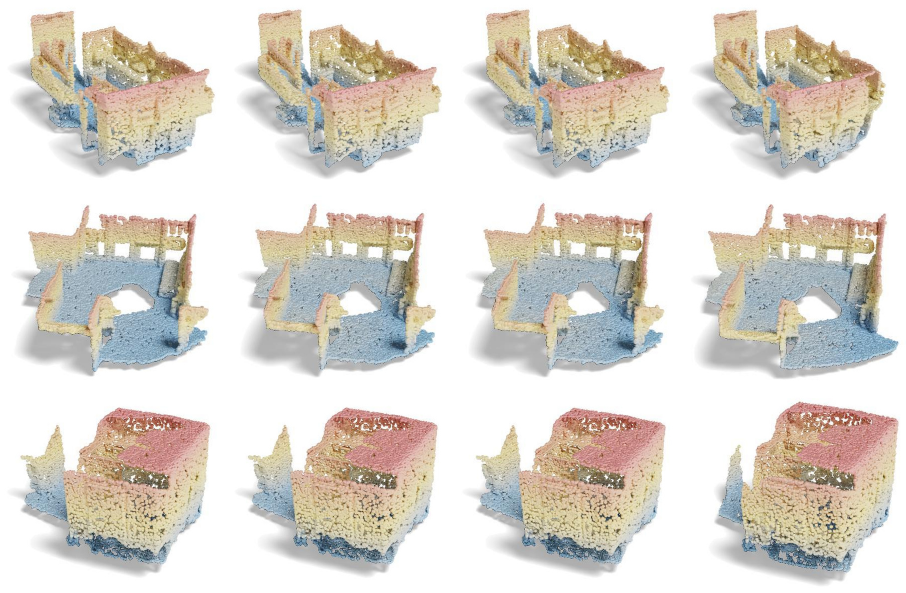}
  \caption{ScanNet examples: Original, Watermarked, Noise, and Crop. Colors are used only for visualization.}
  \label{fig:qual_scannet_main}
\end{figure}

\begin{figure}[t]
  \centering
  \includegraphics[width=\columnwidth]{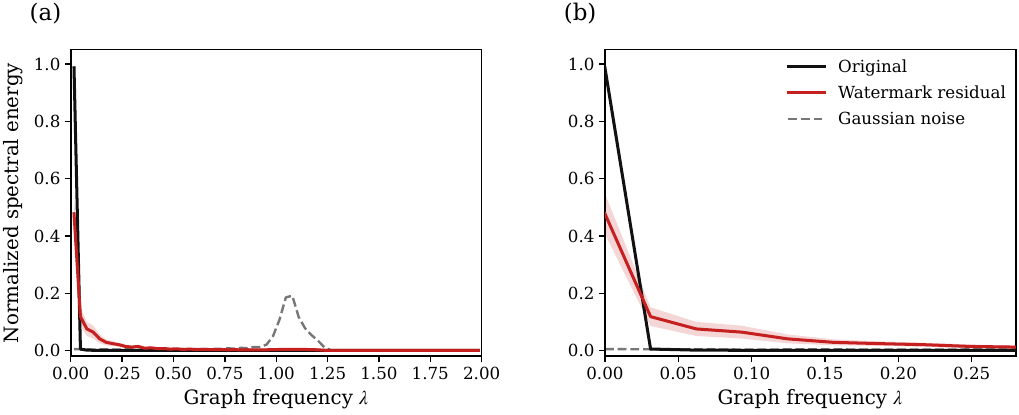}
  \caption{Graph spectral energy distributions. The learned watermark residual concentrates more energy at low frequencies than a Gaussian-noise baseline with matched $\ell_2$ norm.}\label{fig:fre_compare}
  \Description{A spectral analysis figure comparing signal energy across graph Fourier frequencies, highlighting that the learned watermark residual concentrates more strongly in low-frequency bands than matched Gaussian noise.}
\end{figure}

\textbf{Spectral analysis of watermark robustness.}
To understand the observed robustness, we analyze the spectrum of the watermark residual $\Delta\mathbf{x}=\mathbf{x}_w-\mathbf{x}$ using a graph Fourier basis induced by a $k$NN graph on the input points. We compute the normalized graph Laplacian and perform the graph Fourier transform following~\cite{shuman2013emerging,hu2022exploring}. Figure~\ref{fig:fre_compare} compares the spectral energy distributions of the input geometry, the learned watermark residual, and a Gaussian-noise baseline with matched $\ell_2$ norm, averaged over the ShapeNet evaluation set.

The learned residual is concentrated in low-frequency bands, consistent with smoother perturbations and the greater resilience of low-frequency signals to resampling and smoothing~\cite{cox1997secure}. This pattern is consistent with the transformation-aware objective, displacement smoothness, and local covariance matching. It suggests a structural explanation for the observed robustness, although the spectral analysis alone does not establish causality.
\looseness=-1

\begin{figure}[!tbh]
  \centering
  \includegraphics[width=1.0\columnwidth]{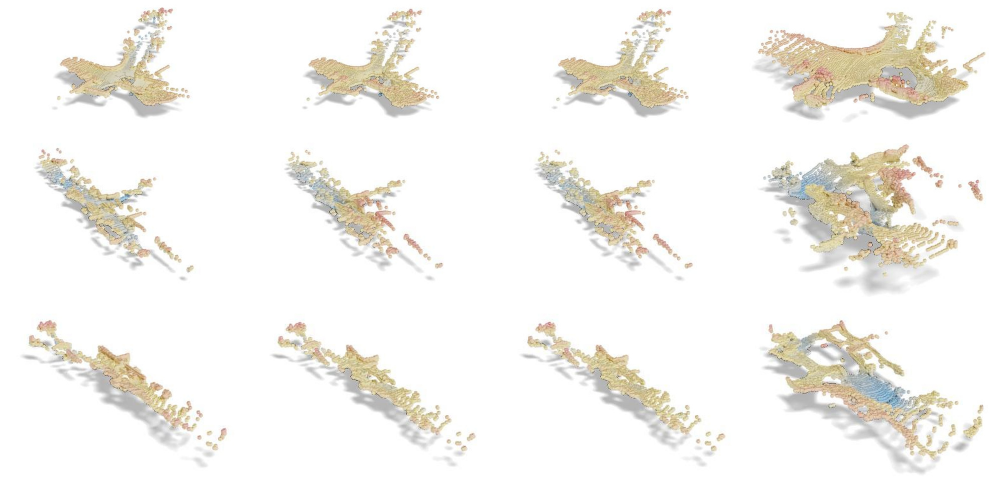}
  \caption{SemanticKITTI examples: Original, Watermarked, Noise, and Crop. Each row shows an outdoor LiDAR scene.}
  \label{fig:qual_kitti_main}
\end{figure}

\subsection{Qualitative Results}
Figure~\ref{fig:qual_compare_main} compares local artifacts on a ScanNet scene. Although the handcrafted methods~\cite{feng2015distance,liu2019novelpctrad2} can achieve competitive global PSNR and CD, their modulation produces spatially coherent, shell-like bands in the matched close-up views. Our learned residual exhibits fewer such artifacts and better preserves the scene's natural appearance.
Figures~\ref{fig:qual_modelnet_main}, \ref{fig:qual_shape_main}, \ref{fig:qual_scannet_main}, and~\ref{fig:qual_kitti_main} show object-level and scene-level examples. Columns are Original, Watermarked, Noise, and Crop; colors are for visualization only and are not model inputs. The clean pairs show geometry preservation after embedding, while the attacked outputs illustrate severe distortions under which decoding remains accurate (Tables~\ref{tab:vertical_compare} and~\ref{tab:main_large_scale}). The images complement these quantitative results rather than measuring recovery accuracy themselves.

\subsection{Ablation Study} We conduct ablations on ModelNet40, ScanNet, and SemanticKITTI, keeping all other hyperparameters fixed.

\textbf{Rotation handling.}
We evaluate progressive alignment by comparing the default model with a variant without alignment. Table~\ref{tab:ablation_rotation_scannet} shows that alignment is especially important under large rotations on ScanNet, where it improves BitAcc from 0.62 to 0.89.

\begin{table}[t]
  \centering
  \caption{Alignment ablation. Clean and Rotation report BitAcc ($\uparrow$); CD ($\downarrow$) and PSNR ($\uparrow$) measure fidelity.}\label{tab:ablation_rotation_scannet}
  \tablefont
  \setlength{\tabcolsep}{10pt}
  \begin{tabular}{@{}llcccc@{}}
    \toprule
    Dataset & Variant & Clean & Rotation & CD & PSNR \\
    \midrule
    ModelNet40 & w/o alignment & 0.99 & 0.82 &  0.001 & 37.0 \\

    ModelNet40 & with alignment & 0.99 & 0.96 &  0.001 & 37.2 \\
    \midrule
    ScanNet & w/o alignment & 0.91 & 0.62 &  8.6e-5 & 46.7 \\

    ScanNet & with alignment & 0.98 & 0.89 &  6.1e-5 & 49.0 \\
    \bottomrule
  \end{tabular}
\end{table}

\textbf{Geometry regularizers and normal consistency.}
We assess the contribution of the structural regularizers: local covariance matching ($\mathcal{L}_{\text{struct}}$) and displacement smoothness ($\mathcal{L}_{\text{smooth}}$). We estimate surface normals using PCA on $k$NN neighborhoods and compute normal consistency (NC) as $\mathrm{NC} = \frac{1}{N}\sum_{i=1}^{N} \left|\langle n_i, n'_i\rangle\right|$. Table~\ref{tab:ablation_geomloss} reports results for ScanNet with a 16-bit payload and ModelNet40 with a 2-bit payload. Adding the regularizers to the baseline Chamfer loss $\mathcal{L}_{\text{cd}}$ reduces geometric error and improves normal consistency in both settings. On ScanNet, for example, NC increases from 0.93 to 0.99.

\begin{table}[t]
  \centering
  \caption{Geometry-loss ablation. Full adds $\mathcal{L}_{\mathrm{struct}}$ and $\mathcal{L}_{\mathrm{smooth}}$ to $\mathcal{L}_{\mathrm{cd}}$; higher BitAcc, NC, and PSNR and lower CD are better.}\label{tab:ablation_geomloss}
  \tablefont
  \setlength{\tabcolsep}{6pt}
  \begin{tabular}{@{}lllcccc@{}}
    \toprule
    Dataset & Payload & Loss setting & BitAcc & CD ($\downarrow$) & NC ($\uparrow$) & PSNR ($\uparrow$)  \\
    \midrule
    ScanNet & 16-bit & $\mathcal{L}_{\text{cd}}$ & 0.95 & 1e-4 & 0.93 & 45.2\\
    ScanNet & 16-bit & Full & 0.98 & 6e-5 & 0.99 & 49.0\\
    \midrule
    ModelNet40 & 2-bit & $\mathcal{L}_{\text{cd}}$ & 0.99 & 2e-3 & 0.90 & 35.4\\
    ModelNet40 & 2-bit & Full & 0.99 & 1e-3 & 0.95 & 37.2\\
    \bottomrule
  \end{tabular}
\end{table}

\textbf{Payload scalability.}
We evaluate payload capacity on SemanticKITTI by increasing the message length from 16 to 128 bits. As reported in Table~\ref{tab:ablation_payload_kitti}, clean BitAcc remains above 0.88 for payloads up to 64 bits, with low geometric distortion. At 64 bits, BitAcc under the listed attacks ranges from 0.820 to 0.879. At 128 bits, decoding accuracy drops further, illustrating the trade-off between payload capacity and robustness.

\begin{table}[t]
  \centering
  \caption{Payload scaling ablation on SemanticKITTI. Higher BitAcc is better; lower CD is better.}\label{tab:ablation_payload_kitti}
  \tablefont
  \setlength{\tabcolsep}{12pt}
  \begin{tabular}{@{}lccccc@{}}
    \toprule
    Payload & Clean & Noise & Crop & Rot. ($[0,\pi]$) & CD \\
    \midrule
    16-bit  & 0.999 & 0.995 & 0.985 & 0.978 & 1.0e-4 \\
    32-bit  & 0.959 & 0.958 & 0.887 & 0.938 & 1.3e-4 \\
    64-bit  & 0.888 & 0.879  & 0.820 & 0.860 & 3.9e-4 \\
    128-bit & 0.739 & 0.729 & 0.691  & 0.718 &  7.2e-4\\
    \bottomrule
  \end{tabular}
\end{table}

\begin{figure}[!tbh]
  \centering
  \includegraphics[width=0.86\columnwidth]{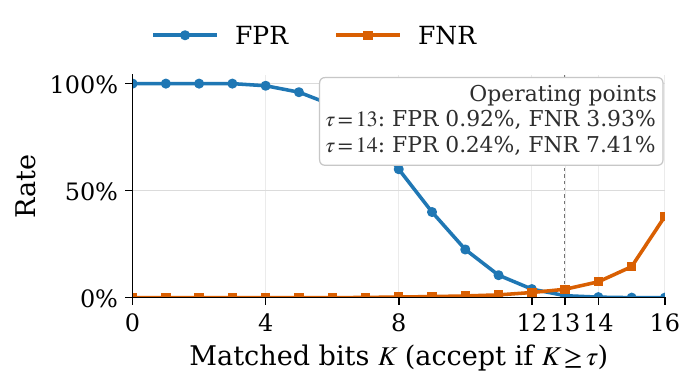}
  \caption{Ownership verification with 16 bits. A higher matched-bit threshold $\tau$ reduces false positives but increases false negatives.}
  \label{fig:threshold_verification_main}
\end{figure}

\subsection{Ownership Verification}
\clubpenalty=10000
Blind decoding of non-watermarked clouds yields chance-level BitAcc: 0.501 on ModelNet40, 0.498 on ShapeNet, 0.503 on ScanNet, and 0.497 on SemanticKITTI. Ownership verification instead requires a binary decision. Let $K$ count the extracted bits matching the owner's $L$-bit message; a claim is accepted if $K\geq\tau$. The false-positive rate (FPR) is the acceptance rate on non-watermarked clouds, and the false-negative rate (FNR) is the rejection rate on watermarked clouds.

Figure~\ref{fig:threshold_verification_main} shows this trade-off for $L=16$. At $\tau=13$, FPR is $0.92\%$ and FNR is $3.93\%$; raising $\tau$ to 14 reduces FPR to $0.24\%$ but increases FNR to $7.41\%$. The threshold therefore depends on the relative costs of false acceptance and rejection. This private-verifier protocol assumes that the owner's message and decoder remain private; it does not provide cryptographic authentication. Keyed message generation, message authentication, collision analysis, and ambiguity attacks remain outside our geometric robustness evaluation.

\section{Conclusion}

We presented a blind watermarking framework for raw 3D point clouds that recovers messages from the observed coordinates alone. The octree encoder--decoder combines transformation-aware training, progressive alignment, and geometry-preserving regularization. Experiments across four object and scene benchmarks demonstrate robust decoding under common geometric distortions, while qualitative comparisons reveal fewer structured local artifacts than handcrafted baselines.

Our evaluation covers common non-adaptive processing rather than cryptographic authentication or an adaptive remover with knowledge of the system. Future work includes adaptive removal, generative reconstruction and completion, diffusion-based denoising, compression, and combined processing chains, as well as higher-capacity and streaming representations.

\clearpage
\bibliographystyle{eg-alpha-doi}
\bibliography{main}

\clearpage
\appendix
\twocolumn[{%
  \refstepcounter{section}\label{app:network}%
  \noindent\textbf{Appendix~\thesection: Network Architecture}\par\medskip
}]

The embedder is a four-level octree U-Net. Its input concatenates the finest-level point features and the $L$-bit message broadcast to occupied nodes; the same message is concatenated again at the bottleneck. Four downsampling stages use 2, 3, 4, and 6 residual blocks, respectively, with channels 64, 128, 128, and 256. Four upsampling stages use two residual blocks each and channels 256, 128, 96, and 96. Features at matching octree depths are joined by skip connections. Interpolation at the original point coordinates and a $96\!\to\!64\!\to\!3$ head predict displacements, which are added to the input points and normalized.

\newpage

The decoder first applies two $3\times3$ alignment heads in sequence, then normalizes the aligned cloud and constructs a new octree. Each head uses a shared pointwise MLP ($3\!\to\!64\!\to\!128\!\to\!1024$), global max pooling, and fully connected layers ($1024\!\to\!512\!\to\!256\!\to\!9$); the predicted matrix is added to the identity. The extractor uses an octree convolution followed by two residual stages with channels 64, 128, and 256, two residual blocks per stage, global pooling, and a $256\!\to\!512\!\to\!L$ head for message logits. Object and scene models share this architecture but are trained separately with octree depths 5 and 7, respectively.

\end{document}